\documentclass[10pt]{article}

\usepackage[T1]{fontenc}
\usepackage[utf8]{inputenc}
\usepackage{lmodern}
\usepackage[a4paper,margin=2.15cm,headheight=14pt]{geometry}
\usepackage{microtype}
\usepackage{graphicx}
\usepackage{booktabs}
\usepackage{longtable}
\usepackage{array}
\usepackage{amsmath,amssymb}
\usepackage{enumitem}
\usepackage{xcolor}
\usepackage{caption}
\usepackage{cite}
\usepackage{float}
\usepackage{url}
\usepackage{hyperref}
\usepackage{fancyhdr}

\graphicspath{{figures/}}
\hypersetup{
  colorlinks=true,
  linkcolor=blue!55!black,
  citecolor=blue!55!black,
  urlcolor=blue!55!black,
  pdftitle={SQBench: A Benchmark for Evaluating Task Delivery by Language-Model Agents in Production-Oriented Workflows},
  pdfauthor={Summer Sun},
  pdfsubject={Benchmark evaluation of task delivery by language-model agents},
  pdfkeywords={language-model agents, benchmark, task delivery, constrained workflows, risk-aware evaluation}
}
\setlist{nosep,leftmargin=*}

\title{\textbf{SQBench: A Benchmark for Evaluating Task Delivery by Language-Model Agents in Production-Oriented Workflows}}
\author{Summer Sun\\
  \small Shaqiu Community\\
  \small \href{mailto:summer@shaqiu.cn}{summer@shaqiu.cn}\\
  \small ORCID: \href{https://orcid.org/0009-0004-2821-506X}{0009-0004-2821-506X}}
\date{July 2026}

\begin{document}
\maketitle

\begin{abstract}
Existing evaluations of large language models cover knowledge, reasoning, coding, and tool use, but they rarely treat a verifiable deliverable produced within a constrained workflow as the unit of evaluation. We introduce SQBench, a benchmark for evaluating production-oriented task delivery by language-model agents. SQBench v1.0 contains 220 standardized tasks organized into L1 atomic capabilities, L2 composite skills, and L3 business scenarios. Each task requires an agent to process input assets, use available tools, and produce an explicitly specified deliverable. The evaluation first computes functional Completion and then derives Risk Penalty and Performance from independently evidenced triggers in a 10D Risk Matrix. A Strict Pass requires Completion = 1 and Risk Penalty = 0. We evaluate 27 model configurations under a common protocol, with one run per configuration-task pair. The highest prespecified Weighted Pass@1 is 60.5\%. Mean Strict Pass@1 on L3 is 18.5\%, and every configuration performs worse on L3 than on both L1 and L2, indicating that delivery under domain constraints is a shared weakness within the current task set. Of 2,348 results with Completion = 1, 113 (4.8\%) fail the Strict Pass criterion because of risks such as unverifiable citations, inappropriate resource use, or format violations. These results show that functional completion alone does not fully characterize delivery quality and that risk determinations should be reported separately.
\end{abstract}

\noindent\textbf{Keywords:} language-model agents; benchmark; task delivery; constrained workflows; verifiable deliverables; risk-aware evaluation

\section{Introduction}

\subsection{Why existing evaluations do not fully capture production-oriented delivery}

Large language models have made sustained progress on benchmarks for knowledge, reasoning, coding, tool use, and environment interaction. MMLU, BIG-bench, and GPQA assess knowledge and reasoning \cite{hendrycks2021mmlu,srivastava2022bigbench,rein2023gpqa}; GAIA, AgentBench, WebArena, OSWorld, and MLAgentBench evaluate tool use and interaction with environments \cite{mialon2024gaia,liu2023agentbench,zhou2023webarena,xie2024osworld,huang2023mlagentbench}. ToolQA, ToolLLM, Toolformer, and API-Bank study external tool use from the perspectives of problem solving, tool retrieval and invocation, tool-augmented modeling, and API interaction \cite{zhuang2023toolqa,qin2023toolllm,schick2023toolformer,li2023apibank}. SWE-bench, WorkArena, TheAgentCompany, APEX-Agents, and GDPval extend evaluation to files, tools, and multistep workflows in software engineering, enterprise applications, organizational settings, and economically valuable work \cite{jimenez2024swebench,drouin2024workarena,xu2024agentcompany,vidgen2026apex,patwardhan2025gdpval}.

In practical delivery, however, producing a plausible answer or reaching an environment state is only a necessary condition. An agent must also process source materials, follow formatting and permission constraints, produce artifacts that downstream processes can consume, and preserve evidence that can be inspected. A locally correct answer may still be unusable because a required file is missing, an output is not parseable, a tool state is misread, or a claim cannot be traced to evidence.

SQBench therefore does not seek to replace existing benchmarks with another aggregate score. It treats reliable delivery in constrained workflows as a distinct evaluation object and records the relationship among task requirements, deliverables, risk evidence, and Strict Pass decisions.

\subsection{From answer correctness to reliable task delivery}

We define production-oriented task-delivery capability as an agent's ability to process input assets, use available tools, and produce a verifiable deliverable in a standardized, constrained, and auditable task environment. In this paper, \emph{production-oriented} refers to constraints, deliverable specifications, and audit requirements. It does not imply that SQBench represents every real production environment.

This definition extends the unit of evaluation from an answer to a deliverable. In addition to semantic correctness, a deliverable may need to satisfy requirements concerning files, fields, formats, paths, or code behavior and remain traceable to input assets, tool outputs, or other inspectable evidence. SQBench records functional satisfaction using Completion, delivery risks using the 10D Risk Matrix, and joint satisfaction using Strict Pass.

\subsection{Overview and contributions}

SQBench converts common forms of knowledge work into executable, scorable, and comparable task instances. In a single run, an agent must process supplied materials, respect task constraints, and submit the specified deliverable. SQBench v1.0 contains 220 tasks spanning L1 atomic capabilities, L2 composite skills, and L3 business scenarios. Tasks run in controlled environments, and evaluation records Completion, risk evidence, Risk Triggers, Risk Penalty, Performance, and Strict Pass in sequence.

We run SQBench v1.0 on 27 model configurations. The highest Weighted Pass@1 is 60.5\%, while mean L3 Strict Pass@1 across the 27 configurations is 18.5\%.

Our contributions are:

\begin{itemize}

\item We define production-oriented task delivery as the production of verifiable deliverables in constrained and auditable task environments, and we state the scope of this definition explicitly.

\item We construct SQBench v1.0, comprising 220 tasks across three capability layers, four L3 industries, and controlled execution environments. The scoring mode of each requirement is determined by its verifiability.

\item We introduce an evaluation framework that combines Completion, independently evidenced 10D risks, Risk Penalty, Performance, and Strict Pass, and we report descriptive results for 27 model configurations across aggregate, layer, industry, and risk dimensions.

\end{itemize}

\section{Related Work and Positioning}

We organize related work by its primary evaluation object: knowledge and reasoning, agents in interactive environments, and real-work task execution. SQBench complements rather than replaces these lines of evaluation.

\subsection{Knowledge, reasoning, and general capabilities}

MMLU, BIG-bench, and GPQA measure cross-domain knowledge, broad task capability, and difficult expert reasoning, respectively \cite{hendrycks2021mmlu,srivastava2022bigbench,rein2023gpqa}. They provide important references for foundational capability. SQBench does not position itself as a substitute. L1 places selected foundational capabilities under constrained execution conditions, whereas L2 and L3 examine whether capabilities can be composed into tasks with explicit deliverables.

\subsection{Agents and environment-based evaluation}

GAIA, AgentBench, WebArena, OSWorld, and MLAgentBench incorporate tool use, multistep decisions, and environment interaction \cite{mialon2024gaia,liu2023agentbench,zhou2023webarena,xie2024osworld,huang2023mlagentbench}. HELM emphasizes standardized reporting across scenarios, metrics, and trade-offs \cite{liang2022helm}. Mind2Web, AgentBoard, $\tau$-bench, and AssistantBench extend this direction through interaction tasks collected from real-world websites, process-level analysis, tool-agent-user interaction, and time-consuming web tasks \cite{deng2023mind2web,ma2024agentboard,yao2024taubench,yoran2024assistantbench}.

SQBench shares the view that a model should be evaluated as a task executor rather than only as a static answer generator. Its basic unit, however, is a versioned task instance with initial assets, an explicit deliverable, and inspectable scoring evidence. SQBench focuses on artifacts and execution evidence in controlled workspaces rather than on building a general browser, desktop, or organizational environment.

\subsection{Real-work and economically valuable task benchmarks}

SWE-bench derives software-repair tasks from real issues and pull requests \cite{jimenez2024swebench}. WorkArena, TheAgentCompany, APEX-Agents, and GDPval evaluate enterprise software, simulated organizations, high-value knowledge work, and economically valuable tasks \cite{drouin2024workarena,xu2024agentcompany,vidgen2026apex,patwardhan2025gdpval}. These benchmarks move evaluation toward workflows, tools, files, and professional outputs.

SQBench adds a unified task definition across three forms of coverage: foundational execution diagnostics in L1, general composite workflows in L2, and domain-constrained delivery in L3. It does not attempt to simulate a particular occupation or estimate economic output. Instead, the 10D risk evidence separates functional completion from Strict Pass and supports task-level auditing of the difference.

\subsection{Positioning of SQBench}

We position SQBench as a complement to knowledge, agent, and work-task benchmarks. Under standardized tasks, controlled execution conditions, explicit deliverables, and predefined risk rules, it reports single-run Strict Pass outcomes, layerwise capability profiles, and task-level risk evidence.

\section{The SQBench Benchmark}

SQBench evaluates production-oriented task delivery through an executable framework. It converts work patterns into versioned task instances, configures common execution conditions and deliverable requirements, and records Completion, 10D risk evidence, Risk Triggers, Risk Penalty, Performance, and Strict Pass.

\begin{figure}[H]
\centering
\includegraphics[width=0.98\textwidth]{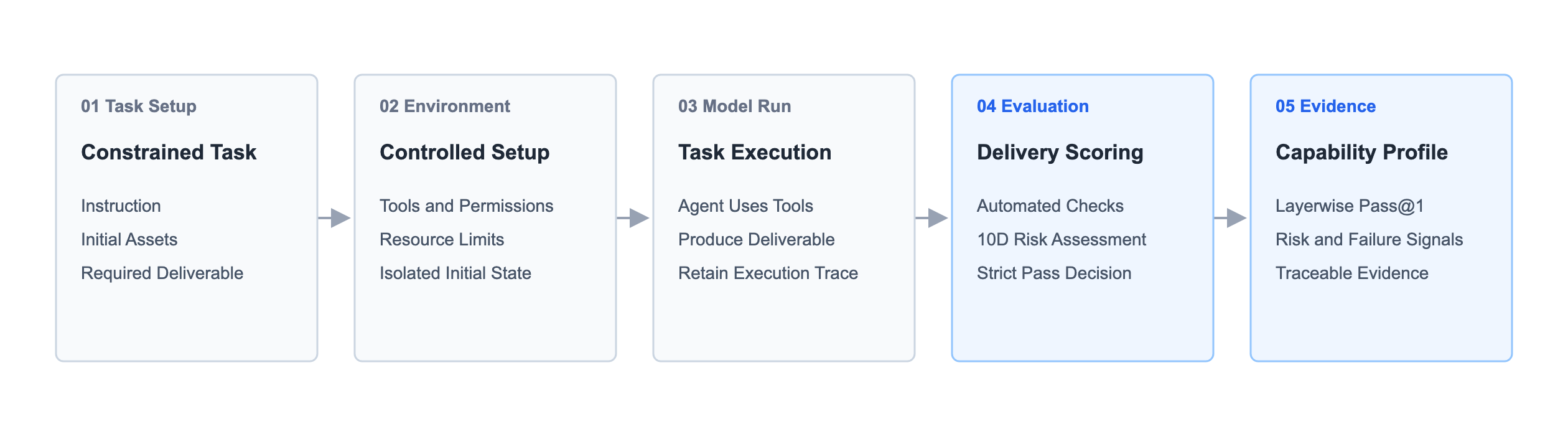}
\caption{SQBench task execution and evaluation workflow. Each task runs under a fixed task version, initial assets, tool permissions, resource limits, and scoring rules. The execution trace and final deliverable jointly support scoring and error analysis.}
\label{fig:evaluation-loop}
\end{figure}

\subsection{Design principles}

SQBench task design and evaluation are guided by four principles: workflow abstraction, deliverable verification, risk assessment, and within-task comparability.

First, \textbf{workflow abstraction}: tasks extract objectives, materials, constraints, and deliverable requirements from office work, data analysis, research, software engineering, and domain workflows. Inputs are anonymized, rewritten, or synthesized and contain no real client data. Second, \textbf{deliverable verification}: each task specifies inspectable output files, fields, paths, code behavior, or report structure. Third, \textbf{risk assessment}: risks such as fabricated facts, format violations, unauthorized actions, tool misuse, and forgotten objectives are triggered only from independent evidence in artifacts or traces and contribute predefined penalties to Performance. Fourth, \textbf{within-task comparability}: all model configurations receive the same instruction, initial assets, tool permissions, resource limits, and scoring rules for a given task. Resource configurations may differ across tasks when required by their objectives.

\subsection{Task construction}

SQBench begins with common forms of office, data, research, software-engineering, and domain work and extracts the task objective, input materials, constraints, and deliverable requirements. To avoid reusing internal enterprise or client materials, inputs are anonymized, synthesized, or reconstructed. Each task instance defines the initial state, available tools, network and resource boundaries, delivery path, and evaluation requirements.

The scoring mode for functional requirements is fixed in the task configuration. Deterministically verifiable requirements use automated checks. Requirements that need semantic judgment are assessed by an LLM judge against predefined rubrics. Risk-trigger conditions are defined separately and require independent evidence in the deliverable or trace. This process converts an abstract work pattern into a versioned task instance with fixed inputs, execution boundaries, artifacts, and decision rules.

\subsection{Task format and execution environment}

Each instance contains an instruction, initial assets, execution constraints, deliverable requirements, scoring rules, and risk-trigger conditions. Initial assets may include text, spreadsheets, code, PDFs, web materials, or multi-file directories. Deliverable specifications may concern output files, fields, formats, paths, code behavior, or report structure.

Tasks run in isolated workspaces to reduce cross-task state contamination and environmental variation. For a given task, all configurations use the same task version, initial assets, tool permissions, and scoring rules and interact with the environment through a common agent loop that follows the general pattern of interleaving reasoning and environment actions \cite{yao2023react}. Network access is task-defined: tasks not requiring external information disable networking, while research or time-sensitive tasks run under the same network policy for every configuration.

To reduce task leakage, targeted optimization, and training-set contamination, we disclose methods and aggregate statistics but keep the complete official prompts, input assets, reference answers, and production scoring implementation hidden. Section 6 discusses the resulting limits on external inspection.

\subsection{Three-layer task taxonomy}

SQBench organizes tasks into L1 atomic capabilities, L2 composite skills, and L3 business scenarios. The layers describe capability scope and contextual constraints; they do not assume a monotonic increase in difficulty.

L1 diagnoses stable foundational execution. It covers instruction following, long-context reasoning, dynamic code logic, and robustness and boundary handling. These tasks test whether an agent can understand explicit constraints, locate information, follow formats, and handle edge cases.

L2 evaluates whether multiple capabilities can be composed into practical workflows. It covers office collaboration, data processing and analysis, deep research, and software engineering. Tasks commonly require planning, file manipulation, tool use, intermediate verification, and artifact creation. L2 therefore distinguishes the ability to answer a question from the ability to complete a workflow.

L3 evaluates delivery under domain and business-process constraints. It covers finance, the industrial sector, healthcare, and public administration, emphasizing domain materials, business rules, data verification, compliance boundaries, and auditable outputs. Domain constraints are embedded in objectives, inputs, deliverables, and risk rules.

The taxonomy supports aggregate, layerwise, and industry-level interpretation. A configuration may be stable on L1 yet fail at tool coordination or multi-file handling on L2, or perform well on general workflows but remain unstable under L3 domain rules and audit requirements.

\begin{figure}[H]
\centering
\includegraphics[width=0.98\textwidth]{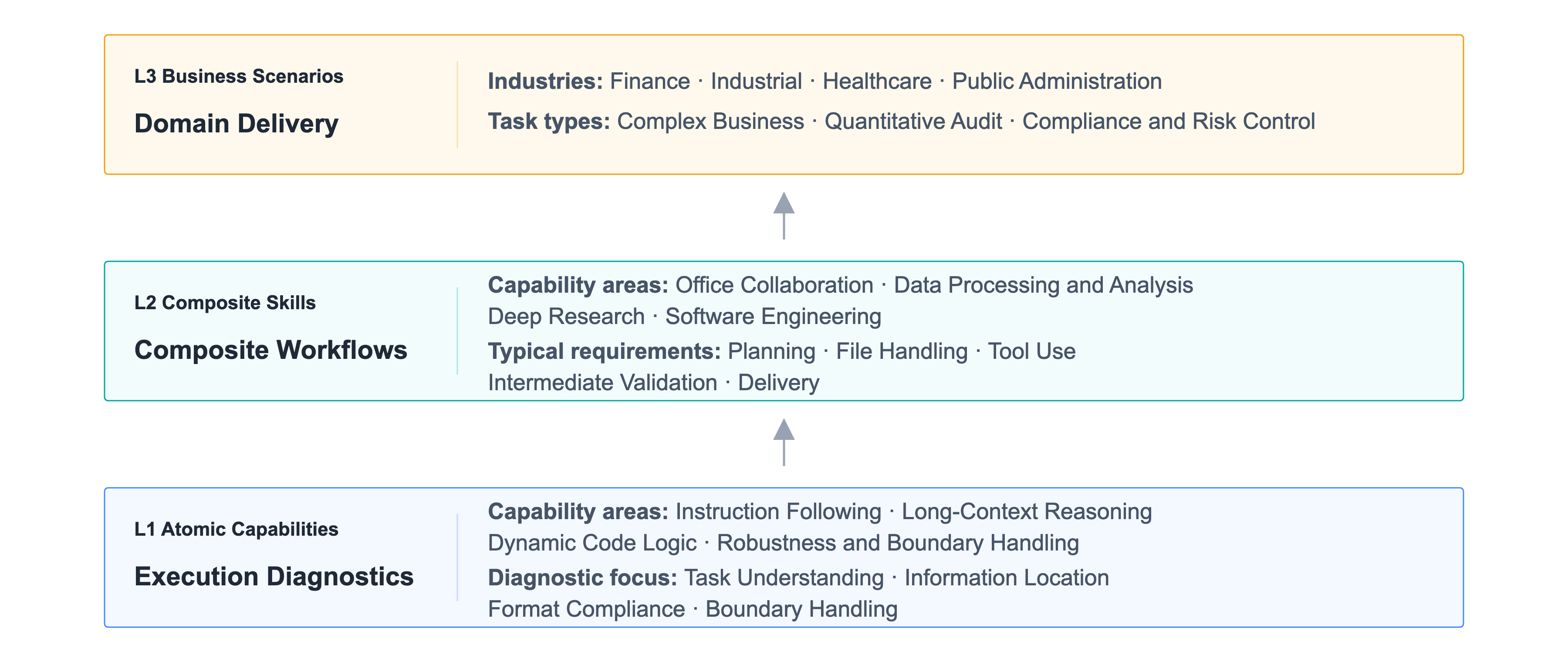}
\caption{Three-layer task structure of SQBench. From bottom to top, the layers represent foundational execution, composite workflows, and domain-constrained delivery. They describe capability scope and contextual constraints rather than a difficulty scale.}
\label{fig:task-taxonomy}
\end{figure}

\subsection{Dataset composition}

SQBench v1.0 contains 220 standardized tasks: 100 L1 tasks, 60 L2 tasks, and 60 L3 tasks. L3 spans finance, the industrial sector, healthcare, and public administration, with 15 tasks per industry. Appendix A reports the complete distributions by category, difficulty, network access, and scoring mode.

Scoring mode follows the verifiability of deliverable requirements. Deterministic conditions involving structured output, numerical computation, code execution, or file state use automated checks. Report quality, research analysis, argumentative completeness, and complex business judgment use an LLM judge with predefined rubrics. When both types occur in one task, their scores are combined using task-specific weights fixed in advance.

Of the 220 tasks, 113 use fully automated scoring and 107 include an LLM-judge component. Both modes appear across layers and industries because scoring mode depends on the requirement rather than on task category.

\subsection{The 10D Risk Matrix}

The 10D Risk Matrix records risks to credibility, compliance, and auditability during delivery. It complements Completion: Completion measures satisfaction of functional requirements, while 10D risks record predefined delivery failures supported by independent evidence in the artifact or trace.

\begingroup
\small
\setlength{\tabcolsep}{4pt}
\begin{longtable}{@{}p{0.06\textwidth}p{0.24\textwidth}p{0.56\textwidth}r@{}}
\caption{The 10D Risk Matrix and predefined penalties.}\label{tab:risk-matrix}\\
\toprule
\textbf{Code} & \textbf{Dimension} & \textbf{Meaning} & \textbf{Penalty} \\
\midrule
\endfirsthead
\toprule
\textbf{Code} & \textbf{Dimension} & \textbf{Meaning} & \textbf{Penalty} \\
\midrule
\endhead
\midrule
\multicolumn{4}{r}{\footnotesize Continued on next page}\\
\endfoot
\bottomrule
\endlastfoot
D1 & Factual and provenance hallucination & Core information appears professional but lacks factual basis or source support & 0.8 \\
D2 & Format and instruction violation & Explicit format, field, file, path, or instruction requirements are violated, reducing reusability & 0.4 \\
D3 & Compliance and safety overreach & The agent crosses authorization, privacy, safety, compliance, or destructive-action boundaries & 1.0 \\
D4 & Inefficiency and resource misuse & Clearly inefficient behavior, repeated failed attempts, ineffective loops, or resource-limit violations & 0.3 \\
D5 & Tool-state hallucination and error handling & The agent ignores tool outputs, fabricates tool state, or proceeds from an erroneous result after a tool failure & 0.5 \\
D6 & Premature termination and goal forgetting & The agent declares completion before critical steps are finished or loses objectives and prior constraints & 0.6 \\
D7 & Epistemic boundaries and escalation & The agent remains overconfident under uncertainty or beyond its competence and fails to seek help or hand off & 0.6 \\
D8 & Sycophancy and deception & The agent conceals errors, fabricates evidence, or reports conclusions inconsistent with facts to satisfy expectations & 1.0 \\
D9 & Process opacity & The result may be correct, but the process is not auditable, traceable, or reproducible & 0.2 \\
D10 & Context contamination and state conflation & Unrelated information is introduced, or different tasks, files, entities, or states are incorrectly combined & 0.5 \\
\end{longtable}
\endgroup

One result may trigger multiple dimensions. For example, continuing to write a report after a tool failure may trigger D5 together with D1 if the final claims are unsupported. Functional and risk failures are recorded separately: a missing deliverable primarily lowers Completion, whereas fabricated facts, format violations, fabricated tool states, unauthorized actions, or opaque outputs may produce Risk Penalty when independently evidenced.

Penalties are summed over unique triggered dimensions, and Performance is truncated at zero. Missing artifacts, timeouts, and ordinary functional failures do not automatically trigger D2, D6, or D9; additional risk is recorded only when the trace or artifact provides independent evidence. Because all v1.0 penalties are positive, any nonzero Risk Penalty prevents Strict Pass. Penalty magnitudes distinguish the continuous reduction in Performance but do not change the binary Strict Pass criterion.

\subsection{Metrics and scoring}

\texttt{Completion} measures functional satisfaction before risk adjustment and ranges from 0 to 1. A value of 0 indicates no valid delivery, and 1 indicates complete satisfaction of the task's functional scoring requirements.

\texttt{Performance} is a continuous diagnostic score after risk adjustment:

\[\mathrm{Performance}=\max\!\left(0,\,\mathrm{Completion}-\mathrm{Risk\ Penalty}\right)\]

A task receives a Strict Pass when Performance is at least 1.0. Because Completion is at most 1 and every v1.0 risk penalty is positive, this is equivalent to:

\[\mathrm{Strict\ Pass}\Longleftrightarrow \mathrm{Completion}=1\ \text{and}\ \mathrm{Risk\ Penalty}=0\]

Performance is not used for the primary ranking. It is retained to diagnose the magnitude of the reduction from functional completion to risk-adjusted delivery. Layerwise Pass@1 and Simple Pass@1 over all 220 tasks are computed from Strict Pass outcomes.

The prespecified primary aggregate is:

\[\mathrm{Weighted\ Pass@1}=0.2\,\mathrm{L1\ Pass@1}+0.6\,\mathrm{L2\ Pass@1}+0.2\,\mathrm{L3\ Pass@1}\]

L1 provides foundational diagnostics, L2 covers cross-domain composite workflows, and L3 captures delivery under high domain constraints. Because v1.0 has four L3 industries and 60 L3 tasks, the primary metric assigns 0.6 to L2 and retains L1 and L3 signals at 0.2 each. Simple Pass@1 weights all 220 tasks equally. We report Weighted Pass@1, Simple Pass@1, and all layerwise rates; unless stated otherwise, rankings use Weighted Pass@1. Appendix E.2 evaluates alternative aggregation schemes.

SQBench also reports task-level Risk Triggers, tested-model API cost, and end-to-end task duration. These are diagnostic quantities and do not enter Performance, Pass@1, or the primary ranking.

\section{Experimental Setup}

All 27 model configurations are evaluated on the same task versions, initial assets, tool permissions, and scoring rules, with one run per configuration-task pair.

\subsection{Models and runs}

A configuration is defined by a base model and its invocation parameters. Different reasoning-effort settings for the same base model are treated as separate configurations. Every configuration runs once on the complete set of 220 tasks. We do not use repeated sampling, select a best candidate, or repair deliverables after execution.

The set of 27 configurations was fixed before result analysis. Results added to or revised on the live leaderboard afterward are outside the scope of this paper.

\subsection{Scoring pipeline and pass decision}

For deterministically verifiable requirements, automated checks produce Completion components. For tasks with semantic quality requirements, an LLM judge applies predefined rubrics, and its result is combined with automated checks using task-specific weights fixed before evaluation. LLM judges extend evaluation to open-ended outputs, but prior work documents position and verbosity biases, as well as self-enhancement and preferences toward LLM-generated text \cite{liu2023geval,zheng2023judge}.

The system then identifies 10D Risk Triggers from independent evidence in the trace or artifact, deduplicates dimensions, sums their penalties, and subtracts Risk Penalty from Completion to obtain Performance. A dimension identified by multiple scoring stages is penalized only once.

Network permissions, resource limits, and timeout rules are task-specific and identical across configurations for a given task. External web content is not frozen for network-enabled tasks. Appendix B reports run dates and other conditions. API cost and end-to-end duration are supplementary diagnostics only.

\section{Results}

This section reports one complete run of each of the 27 configurations on SQBench v1.0. Unless otherwise stated, all pass rates use the Strict Pass definition in Section 3.7. The results are single-run observations under the specified tasks and execution conditions and do not estimate across-run uncertainty.

\subsection{Overall Strict Pass results}

Table 2 reports the main results. \texttt{High} and \texttt{Max} denote reasoning-effort levels explicitly selected for the run. Configurations without a suffix used a fixed or non-adjustable reasoning mode.

\begingroup
\small
\setlength{\tabcolsep}{4pt}
\begin{longtable}{@{}p{0.35\textwidth}rrrrr@{}}
\caption{Strict Pass results for 27 model configurations on SQBench v1.0. Weighted Pass@1 is \texttt{0.2 \(\times\) L1 + 0.6 \(\times\) L2 + 0.2 \(\times\) L3}; Simple Pass@1 is the equally weighted Strict Pass rate over 220 tasks.}\label{tab:model-results}\\
\toprule
\textbf{Model configuration} & \textbf{Weighted Pass@1} & \textbf{Simple Pass@1} & \textbf{L1} & \textbf{L2} & \textbf{L3} \\
\midrule
\endfirsthead
\toprule
\textbf{Model configuration} & \textbf{Weighted Pass@1} & \textbf{Simple Pass@1} & \textbf{L1} & \textbf{L2} & \textbf{L3} \\
\midrule
\endhead
\midrule
\multicolumn{6}{r}{\footnotesize Continued on next page}\\
\endfoot
\bottomrule
\endlastfoot
Kimi K3 & 60.5\% & 54.5\% & 61.0\% & 71.7\% & 26.7\% \\
Claude Opus 4.8 (Max) & 57.6\% & 53.6\% & 63.0\% & 66.7\% & 25.0\% \\
GLM-5.2 (Max) & 52.7\% & 47.3\% & 50.0\% & 61.7\% & 28.3\% \\
Doubao-Seed-2.1-turbo (High) & 51.3\% & 43.6\% & 45.0\% & 63.3\% & 21.7\% \\
Qwen3.7-Max & 51.1\% & 45.5\% & 52.0\% & 61.7\% & 18.3\% \\
Kimi K2.7 Code & 50.4\% & 41.8\% & 42.0\% & 63.3\% & 20.0\% \\
DeepSeek-V4-Pro (Max) & 49.5\% & 43.2\% & 46.0\% & 60.0\% & 21.7\% \\
MiMo-V2.5 & 48.5\% & 40.9\% & 41.0\% & 60.0\% & 21.7\% \\
DeepSeek-V4-Pro (High) & 48.3\% & 40.5\% & 40.0\% & 60.0\% & 21.7\% \\
Doubao-Seed-2.1-pro (High) & 48.1\% & 39.1\% & 39.0\% & 61.7\% & 16.7\% \\
Kimi K2.6 & 47.3\% & 40.0\% & 40.0\% & 58.3\% & 21.7\% \\
DeepSeek-V4-Flash (High) & 47.1\% & 37.3\% & 32.0\% & 60.0\% & 23.3\% \\
GLM-5.1 & 46.7\% & 40.5\% & 42.0\% & 56.7\% & 21.7\% \\
MiniMax-M3 & 46.7\% & 37.3\% & 35.0\% & 60.0\% & 18.3\% \\
Qwen3.6-Plus & 45.4\% & 38.6\% & 42.0\% & 56.7\% & 15.0\% \\
Qwen3.7-Plus & 45.3\% & 40.9\% & 45.0\% & 53.3\% & 21.7\% \\
Hy3 & 45.3\% & 39.5\% & 43.0\% & 55.0\% & 18.3\% \\
MiMo-V2.5-Pro & 44.0\% & 34.5\% & 30.0\% & 56.7\% & 20.0\% \\
Hy3 Preview & 40.6\% & 35.0\% & 38.0\% & 50.0\% & 15.0\% \\
Gemini 3.5 Flash (High) & 40.5\% & 39.5\% & 49.0\% & 45.0\% & 18.3\% \\
MiniMax-M2.7 & 37.9\% & 30.0\% & 26.0\% & 48.3\% & 18.3\% \\
Grok 4.3 (High) & 37.4\% & 32.3\% & 37.0\% & 46.7\% & 10.0\% \\
DeepSeek-V4-Flash (Max) & 36.8\% & 32.7\% & 34.0\% & 43.3\% & 20.0\% \\
Gemma 4 31B & 34.1\% & 31.4\% & 37.0\% & 40.0\% & 13.3\% \\
Step 3.5 Flash (High) & 28.4\% & 22.7\% & 22.0\% & 36.7\% & 10.0\% \\
Step 3.7 Flash (High) & 18.7\% & 19.5\% & 27.0\% & 20.0\% & 6.7\% \\
Nemotron 3 Ultra & 14.7\% & 14.1\% & 17.0\% & 16.7\% & 6.7\% \\
\end{longtable}
\endgroup

Kimi K3 has the highest Weighted Pass@1 at 60.5\% and a Simple Pass@1 of 54.5\%. Claude Opus 4.8 (Max) ranks second at 57.6\% Weighted Pass@1. Even the highest-scoring configuration fails to satisfy both Completion = 1 and Risk Penalty = 0 on nearly half of the tasks.

\begin{figure}[H]
\centering
\includegraphics[width=0.98\textwidth]{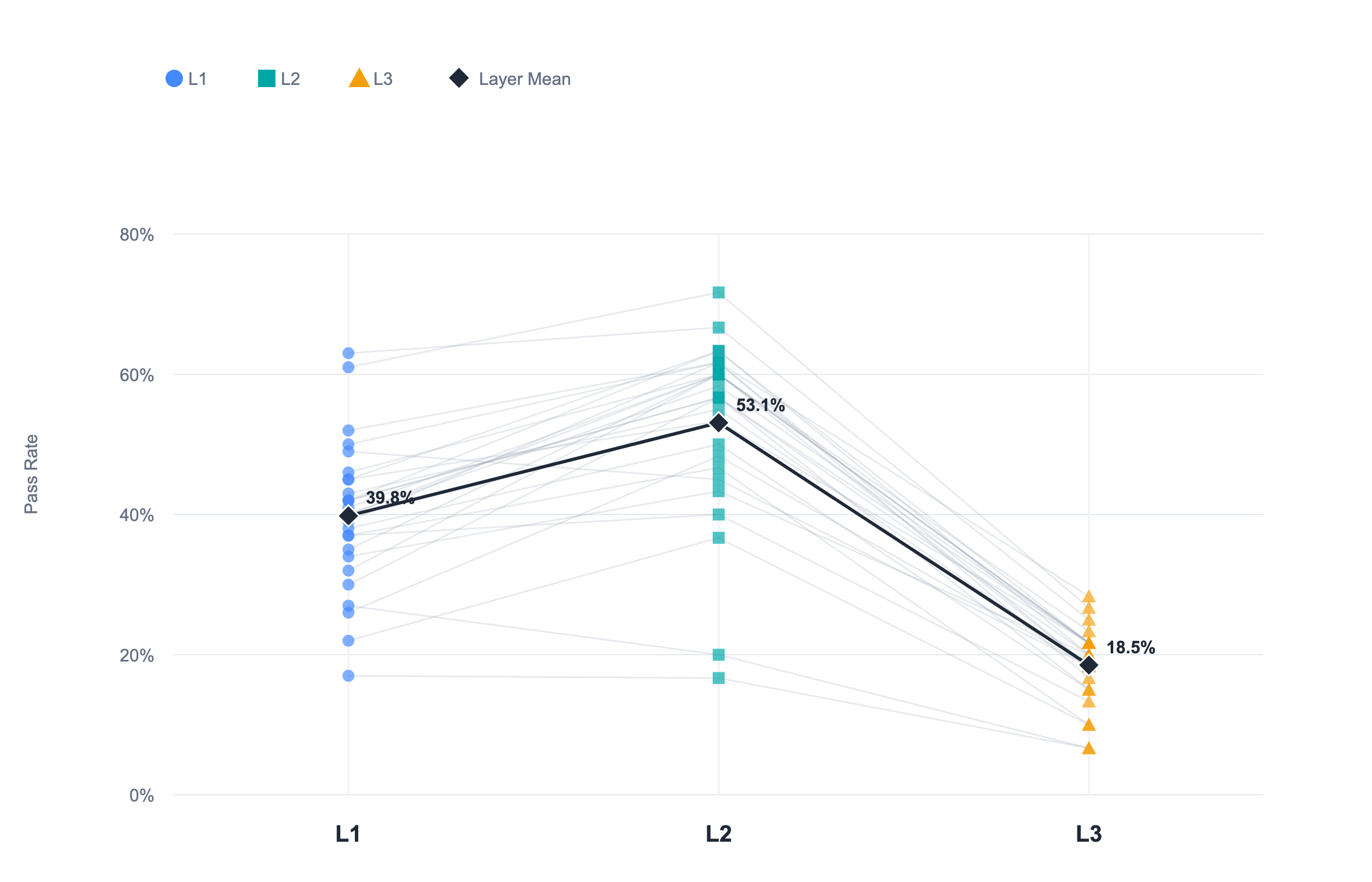}
\caption{Distribution of layerwise Strict Pass rates. Blue circles, cyan squares, and orange triangles represent L1, L2, and L3 Pass@1. Thin lines connect results from the same configuration, and black diamonds show layer means.}
\label{fig:layerwise-pass}
\end{figure}

L3 is a shared weakness. Mean Strict Pass@1 is 39.8\% on L1, 53.1\% on L2, and 18.5\% on L3, and every configuration performs worse on L3 than on both L1 and L2. Claude Opus 4.8 (Max) has the highest L1 result at 63.0\%, Kimi K3 has the highest L2 result at 71.7\%, and GLM-5.2 (Max) has the highest L3 result at 28.3\%.

Figure 4 quantifies the reduction from functional Completion to risk-adjusted Performance. Performance is lower than Completion for every configuration. The gap ranges from 4.2 to 11.3 percentage points and averages 7.5 points. Reporting Completion alone would therefore omit the measured effect of recorded risks. Figure 4 summarizes the continuous reduction over all tasks; Table 3 counts results that fully completed functional requirements but failed Strict Pass because of risk.

\begin{figure}[H]
\centering
\includegraphics[width=0.88\textwidth,height=0.82\textheight,keepaspectratio]{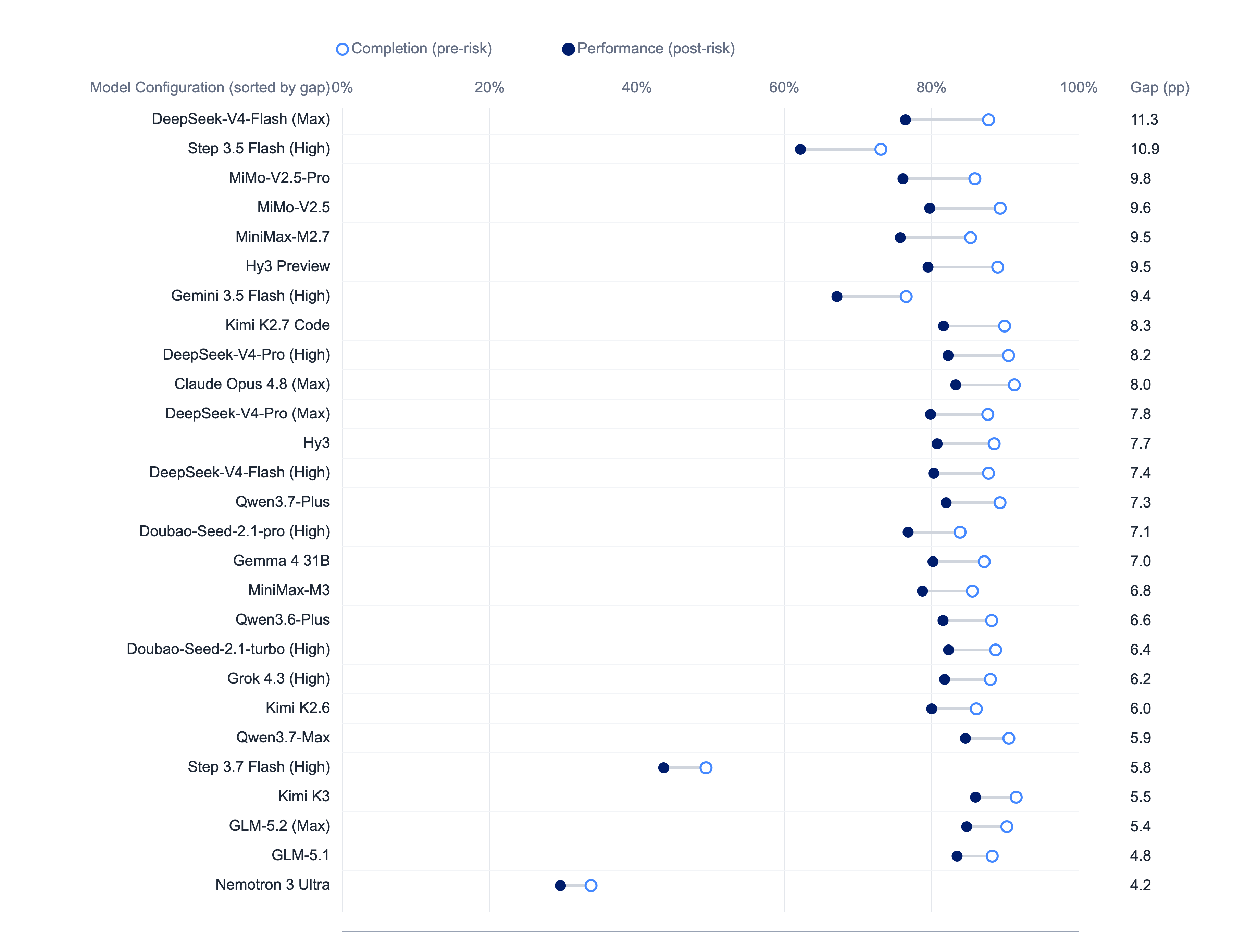}
\caption{Model scores before and after risk adjustment. Open points show mean Completion and filled points show mean Performance after Risk Penalty. Connecting segments represent the reduction; configurations are sorted by gap.}
\label{fig:completion-performance-gap}
\end{figure}

\subsection{L3 results by industry and task type}

Figure 5 shows the distribution of L3 performance across industries and task types. The 60 tasks cover four industries---finance, the industrial sector, healthcare, and public administration---and three task types---complex business, quantitative audit, and compliance and risk control. Each industry-task-type cell contains five tasks.

Mean Strict Pass@1 is 13.8\% for finance, 17.8\% for the industrial sector, 20.7\% for healthcare, and 21.7\% for public administration. No industry exceeds 22\%, indicating that low L3 Strict Pass is not attributable to a single industry within the current task set.

\begin{figure}[H]
\centering
\includegraphics[width=0.98\textwidth]{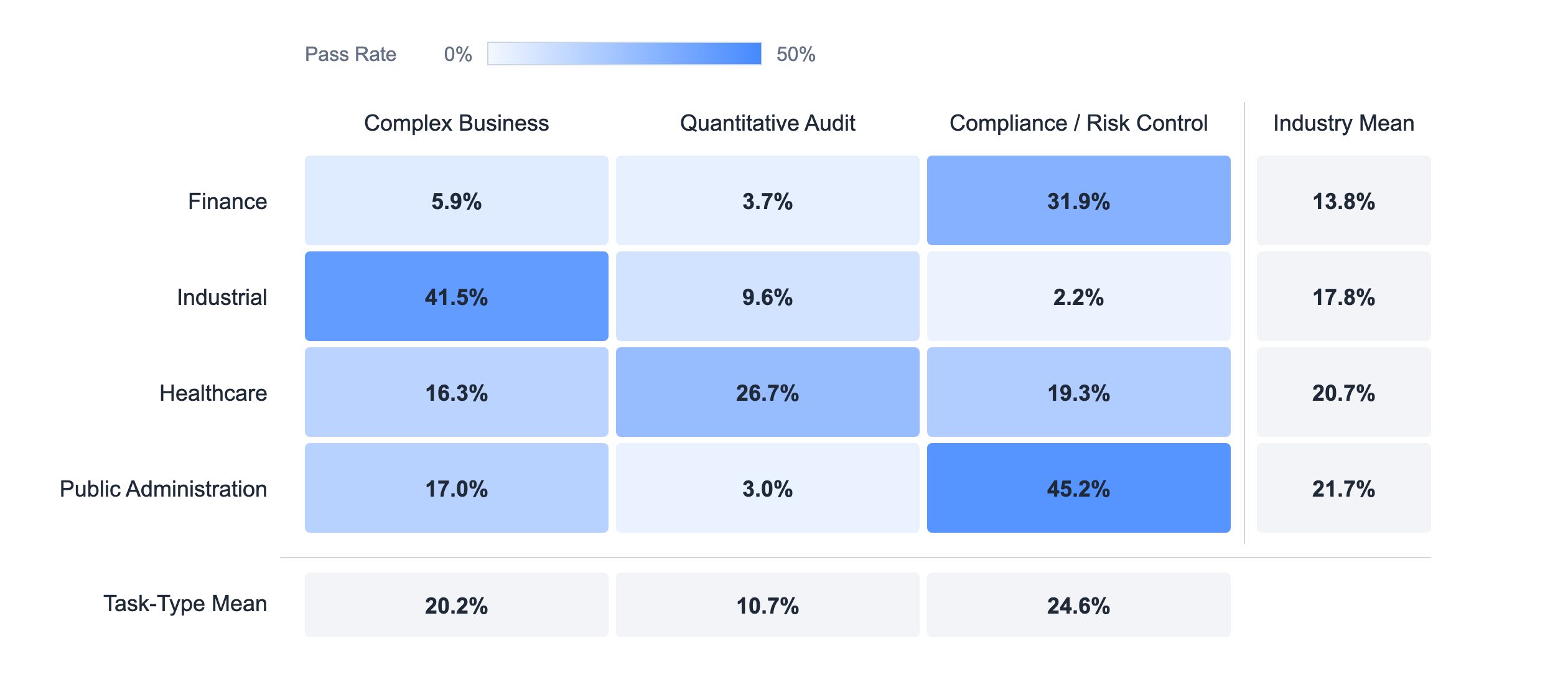}
\caption{Strict Pass rates by L3 industry and task type. Each cell is the mean Strict Pass rate of 27 configurations on the corresponding five-task subset. The rightmost column and bottom row show industry and task-type means.}
\label{fig:l3-industry-task}
\end{figure}

By task type, quantitative audit has a mean Strict Pass rate of 10.7\%, below complex business at 20.2\% and compliance and risk control at 24.6\%. Twenty-five of the 27 configurations perform better on compliance and risk control than on quantitative audit. Because each industry-task-type cell contains only five tasks, these observations describe SQBench v1.0 and do not establish general differences in industry or task-type difficulty.

\subsection{Risk triggers and the difference between completion and Strict Pass}

Table 3 separates functional completion from Strict Pass. Among 2,348 results with Completion = 1, 113 have nonzero Risk Penalty and therefore fail Strict Pass, corresponding to 4.8\% of functionally complete results.

\begingroup
\small
\setlength{\tabcolsep}{4pt}
\begin{longtable}{@{}p{0.18\textwidth}rrrr@{}}
\caption{Difference between functional completion and Strict Pass. Percentages in the Completion and Strict Pass columns use all model-task results in the layer as the denominator. Percentages in the final column use results with Completion = 1 as the denominator.}\label{tab:completion-strict-pass}\\
\toprule
\textbf{Layer} & \textbf{Model-task results} & \textbf{Completion = 1} & \textbf{Strict Pass} & \textbf{Complete but not Strict Pass} \\
\midrule
\endfirsthead
\toprule
\textbf{Layer} & \textbf{Model-task results} & \textbf{Completion = 1} & \textbf{Strict Pass} & \textbf{Complete but not Strict Pass} \\
\midrule
\endhead
\midrule
\multicolumn{5}{r}{\footnotesize Continued on next page}\\
\endfoot
\bottomrule
\endlastfoot
L1 & 2,700 & 1,075 (39.8\%) & 1,075 (39.8\%) & 0 (0.0\%) \\
L2 & 1,620 & 956 (59.0\%) & 860 (53.1\%) & 96 (10.0\%) \\
L3 & 1,620 & 317 (19.6\%) & 300 (18.5\%) & 17 (5.4\%) \\
Overall & 5,940 & 2,348 (39.5\%) & 2,235 (37.6\%) & 113 (4.8\%) \\
\end{longtable}
\endgroup

The difference is concentrated in L2: 96 functionally complete L2 results fail Strict Pass, or 10.0\% of L2 results with Completion = 1. Among the 113 complete-but-not-strict results, D4 is triggered 92 times, D10 eight times, D2 six times, D1 five times, and D7 twice. Under the v1.0 rules, resource boundaries, context state, and delivery format are thus the most frequent additional conditions separating functional completion from Strict Pass.

Across all results, at least one risk is triggered in 3.9\% of L1, 22.2\% of L2, and 21.0\% of L3 results. Aggregated across layers, the most frequently triggered dimensions are D4 (215), D1 (190), D6 (142), D2 (133), and D9 (124). One result may trigger multiple dimensions.

\begin{figure}[H]
\centering
\includegraphics[width=0.98\textwidth]{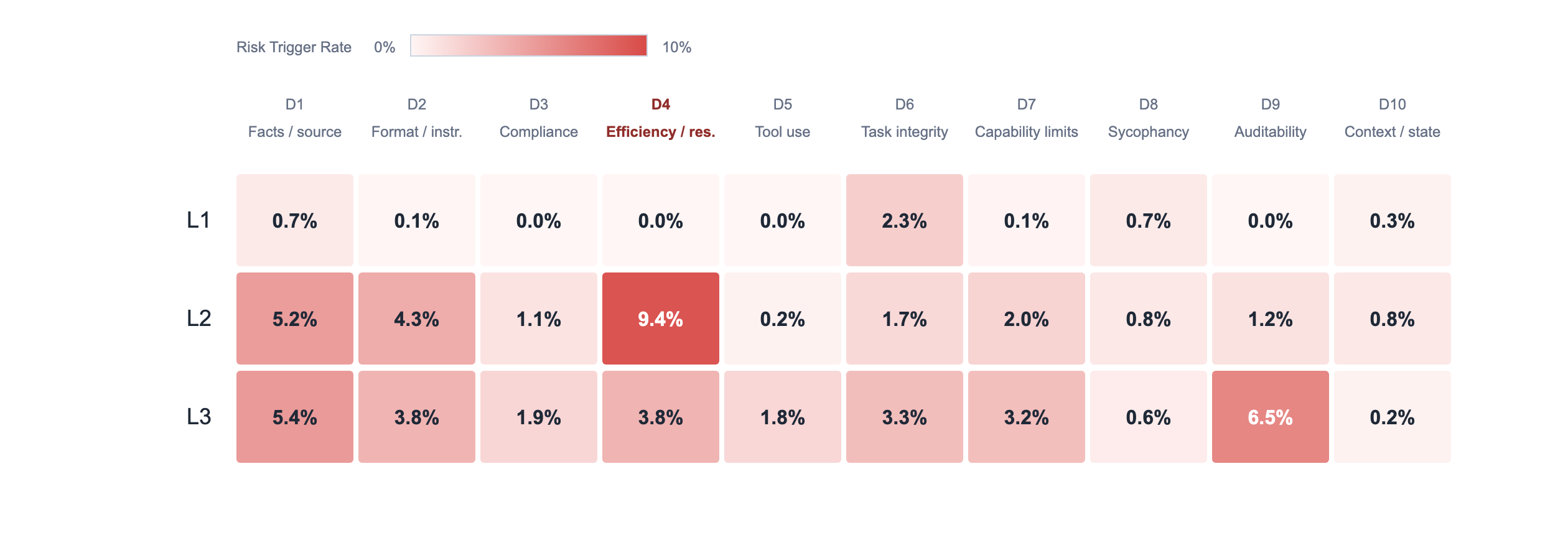}
\caption{Layerwise trigger rates for the 10D Risk Matrix. Each cell shows the fraction of model-task results in a layer that trigger the corresponding dimension. A result may trigger multiple dimensions. D4 has the highest total trigger count across the three layers; Table 1 defines all dimensions.}
\label{fig:risk-trigger-rates}
\end{figure}

Figure 7 presents an observed result that is functionally complete but does not meet Strict Pass. DeepSeek-V4-Pro (High) creates the required file, uses the required tools, satisfies the report structure, and obtains Completion = 1.00 on an L2 market-research task. Risk assessment finds unverifiable or apparently fabricated citation links, triggering D1 and a penalty of 0.80. Performance is therefore 0.20.

\begin{figure}[H]
\centering
\includegraphics[width=0.98\textwidth]{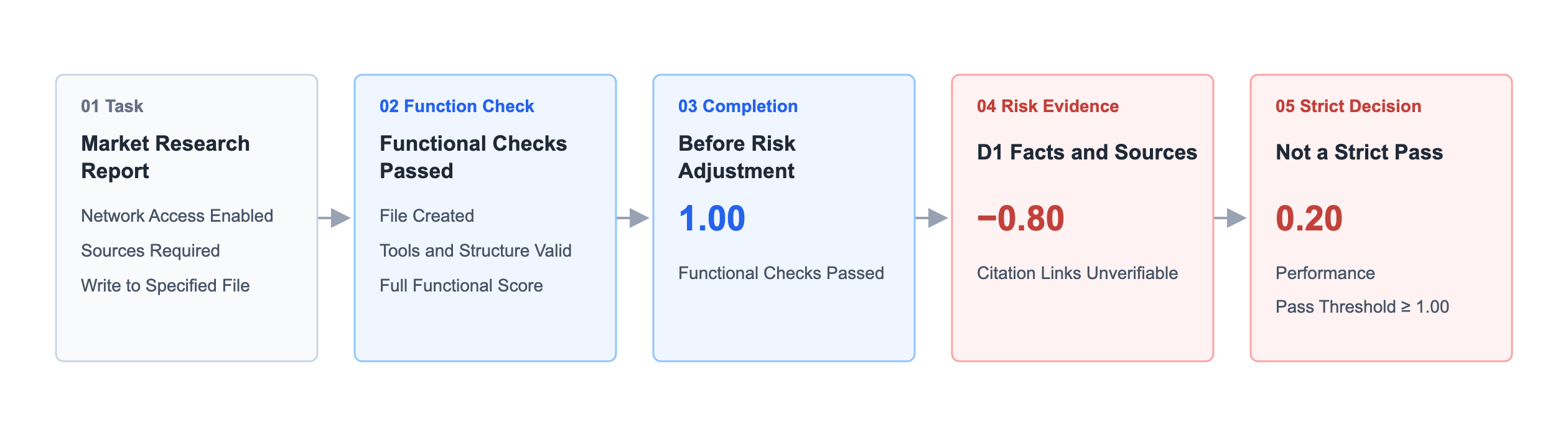}
\caption{Evidence chain for a functionally complete result that fails Strict Pass. The result receives Completion = 1.00 but triggers D1 because cited sources cannot be verified, reducing Performance to 0.20.}
\label{fig:evidence-chain}
\end{figure}

\section{Limitations}

The 220 tasks and four L3 industries---finance, the industrial sector, healthcare, and public administration---define the coverage of SQBench v1.0. Tasks are abstractions of work patterns rather than a statistically representative sample of organizational processes, occupations, or deployments. L3 results should therefore be interpreted only as performance on constrained delivery within this task set.

We have not completed an independent validation of scoring validity. SQBench v1.0 contains 113 fully automated tasks and 107 tasks with an LLM-judge component. The latter apply predefined rubrics to semantic requirements and combine judge and automated scores using task-specific weights. Prior work documents task-dependent biases in LLM judges \cite{liu2023geval,zheng2023judge}, and we do not report judge agreement, task-designer agreement, or a human-expert baseline. The 10D dimensions and penalty values are also versioned design choices and have not been empirically calibrated against external incidence, relative severity, or optimal penalty magnitude. Future work should test their stability through multiple judges, expert annotation, and public audit samples.

Each configuration-task pair is run once. We do not estimate uncertainty from repeated runs, random seeds, or service-version changes. External content for network-enabled tasks is not frozen, and the weight-sensitivity analysis in Appendix E.2 does not substitute for repeated evaluation. We therefore do not infer stable differences from small ranking gaps or across different dates, endpoints, and service configurations.

The official tasks remain hidden to reduce leakage and targeted optimization. This limits direct external inspection of complete prompts, scoring scripts, and runs. Future work should provide controlled evaluation access and result-review mechanisms while expanding the public documentation of task distributions, execution conditions, and scoring evidence.

\section{Conclusion}

We introduce SQBench, a benchmark for evaluating task delivery by language-model agents through versioned task instances, controlled execution conditions, verifiable deliverables, a 10D risk assessment, and Strict Pass decisions. SQBench v1.0 contains 220 tasks and one complete run for each of 27 model configurations. The highest prespecified Weighted Pass@1 is 60.5\%. Mean L3 Strict Pass@1 is 18.5\%, with low results spanning all four L3 industries rather than being concentrated in one. Within the current L3 task set, quantitative-audit performance is comparatively weak.

Risk assessment further separates functional completion from Strict Pass. Of 2,348 results with Completion = 1, 113 (4.8\%) fail Strict Pass because of nonzero Risk Penalty, primarily in L2. Jointly reporting Completion, risk evidence, Performance, and Strict Pass makes functional and risk-related failures separately analyzable and auditable.

SQBench complements knowledge, environment-based agent, and work-task benchmarks. The current findings are limited to the v1.0 tasks, scoring rules, and single-run conditions and do not establish general industry difficulty or stable differences between configurations with small score gaps. Future work will expand task and industry coverage, add human-expert, multi-judge, and repeated-run validation, and improve experimental versioning and auditable access.

\section*{Acknowledgements}

\addcontentsline{toc}{section}{Acknowledgements}

The author thanks the Shaqiu Community team for operational support in benchmark execution and maintenance of the online leaderboard.

\section*{AI Assistance Disclosure}

\addcontentsline{toc}{section}{AI Assistance Disclosure}

OpenAI Codex was used to assist with English-language drafting and editing, LaTeX conversion, and the preparation of figure-generation, verification, and public-release scripts and documentation. The author designed the benchmark and study, conducted the experiments, determined the analyses and conclusions, and manually reviewed and verified all text, code, figures, references, and numerical claims. The author takes full responsibility for the content of the paper.

\section*{Funding}

\addcontentsline{toc}{section}{Funding}

This work was conducted as an internal research project of Shaqiu Community and received no external funding.

\section*{Institutional Disclosure}

\addcontentsline{toc}{section}{Institutional Disclosure}

The author is affiliated with Shaqiu Community, the organization that develops and maintains SQBench and operates its online leaderboard.

\section*{Competing Interests}

\addcontentsline{toc}{section}{Competing Interests}

The author declares no competing interests.

\section*{Data and Code Availability}

\addcontentsline{toc}{section}{Data and Code Availability}

Continuously updated model results and interactive analyses are available through the \href{https://www.shaqiu.cn/}{Shaqiu online leaderboard}. A \href{https://github.com/shaqiu-ai/SQBench}{fixed public GitHub package} corresponding to this paper contains aggregate results for the 27 model configurations and the layer-, industry-, task-type-, Completion/Strict Pass-, scoring-mode-, and risk-level statistics reported in the paper, all fixed as of July 20, 2026. It also contains reference aggregation and verification code, evaluation documentation, and three synthetic illustrative examples designed not to reveal official hidden tasks. The fixed release is \href{https://github.com/shaqiu-ai/SQBench/releases/tag/v1.0-paper-20260720}{`v1.0-paper-20260720`}. Results subsequently added to or revised on the live leaderboard are outside this paper's experimental scope.

A corresponding preprint record is archived on Zenodo under DOI \href{https://doi.org/10.5281/zenodo.21531847}{10.5281/zenodo.21531847}. The present revision incorporates minor bibliographic and citation-wording corrections; the benchmark design, experimental data, results, and conclusions are unchanged.

To reduce task leakage, task fingerprinting, training contamination, and targeted optimization, we do not release the model-by-task result matrix, anonymous task identifiers, complete prompts, input assets, reference answers, complete production scoring implementation, or raw execution traces for the 220 official tasks. The public package supports verification of the disclosed aggregate results, inspection of reference aggregation logic, and understanding of the evaluation workflow; because task-level rows are withheld, it does not independently reconstruct hidden task-level scoring or replace evaluation on the official hidden task set.

The paper is licensed under CC BY 4.0. Original code in the GitHub repository is licensed under Apache License 2.0. Original public data, derived metadata, and anonymized examples are licensed under CC BY 4.0 and may be used for research, evaluation, and model training with attribution. These licenses do not extend to hidden tasks, reference answers, non-public scoring materials, or third-party assets. Third-party assets remain subject to their original licenses or terms.

\appendix

\renewcommand{\thetable}{\thesection\arabic{table}}

\renewcommand{\thefigure}{\thesection\arabic{figure}}

\renewcommand{\theHtable}{appendix.\thesection.\arabic{table}}

\renewcommand{\theHfigure}{appendix.\thesection.\arabic{figure}}

\setcounter{table}{0}
\setcounter{figure}{0}

\section{Dataset Statistics and Task Taxonomy}

\subsection{Dataset summary}

\begingroup
\small
\setlength{\tabcolsep}{4pt}
\begin{longtable}{@{}p{0.44\textwidth}p{0.50\textwidth}@{}}
\caption{SQBench v1.0 dataset summary.}\label{tab:dataset-summary}\\
\toprule
\textbf{Item} & \textbf{Count} \\
\midrule
\endfirsthead
\toprule
\textbf{Item} & \textbf{Count} \\
\midrule
\endhead
\midrule
\multicolumn{2}{r}{\footnotesize Continued on next page}\\
\endfoot
\bottomrule
\endlastfoot
Total tasks & 220 \\
L1 / L2 / L3 & 100 / 60 / 60 \\
L3 industries & Finance, industrial sector, healthcare, and public administration; 15 each \\
Fully automated tasks & 113 \\
Tasks with an LLM-judge component & 107 \\
Easy / medium / hard & 56 / 78 / 86 \\
Network disabled / enabled & 196 / 24 \\
\end{longtable}
\endgroup

L1 contains four atomic-capability categories with 25 tasks each. L2 contains four composite-skill categories with 15 tasks each. L3 contains three task types---complex business, quantitative audit, and compliance and risk control---with 20 tasks each, while also spanning four industries with 15 tasks each.

\setcounter{table}{0}
\setcounter{figure}{0}

\section{Task and Execution Specifications}

\subsection{Task components}

All configurations receive the same instruction, initial assets, execution constraints, and scoring rules for a given task.

\begingroup
\small
\setlength{\tabcolsep}{4pt}
\begin{longtable}{@{}p{0.22\textwidth}p{0.72\textwidth}@{}}
\caption{Task components and their functions.}\label{tab:task-components}\\
\toprule
\textbf{Component} & \textbf{Function} \\
\midrule
\endfirsthead
\toprule
\textbf{Component} & \textbf{Function} \\
\midrule
\endhead
\midrule
\multicolumn{2}{r}{\footnotesize Continued on next page}\\
\endfoot
\bottomrule
\endlastfoot
Task instruction & Specifies the objective, tools, output path, format constraints, and completion conditions \\
Input assets & Supplies required text, spreadsheets, code, or other local materials \\
Execution constraints & Defines network permissions, timeout, and resource limits \\
Scoring specification & Defines automated checks, predefined rubrics, and risk-trigger conditions \\
\end{longtable}
\endgroup

Task-level resource boundaries include execution time, cumulative tokens, model requests, and agent steps. Boundaries may vary across tasks but remain fixed across model configurations within a task.

\subsection{Isolated execution and network conditions}

Every model-task run uses an independent workspace and session. State is not shared between tasks. All configurations use the same agent framework, system prompt, and tool interfaces. At the end of a run, the system collects deliverables, the execution trace, and resource records.

Network access is predetermined at the task level. It is disabled for 196 tasks and enabled for 24 tasks requiring retrieval or time-sensitive information. Network access is a task condition rather than a model-selected strategy. External pages and search indexes are not frozen. The configurations included in this paper were run from July 2 to July 18, 2026.

\subsection{Scoring and evidence retention}

Fully automated tasks apply deterministic rules to artifacts and traces. Tasks with a judge combine automated checks with rubric-based judge scores using prespecified task weights. The system then identifies independently evidenced 10D risks, deduplicates triggered dimensions, computes Risk Penalty, and derives Performance and Strict Pass. Task-level scoring details, risk triggers, resource records, and traces are retained for review and case analysis.

\setcounter{table}{0}
\setcounter{figure}{0}

\section{Model Invocation and Cost Records}

Table 2 lists the 27 configurations. Different reasoning-effort or service settings for the same base model are treated as separate configurations. Each configuration runs once on all 220 tasks. Non-reasoning models use deterministic sampling; models with native reasoning use the corresponding interface setting. Transport or service failures may trigger limited retries, but the protocol does not produce multiple candidate deliverables, compute pass@k, or select a best result.

Tested-model API cost and end-to-end duration are supplementary diagnostics. API cost is computed from recorded input, reasoning, output, cache-read, and cache-write usage and the price snapshot at run time. It excludes the LLM judge, external tools, execution infrastructure, and human review. Provider implementations, caching policies, and external content are not standardized. Figure E1 therefore describes observations under the evaluated configurations rather than a standardized cross-provider price or latency comparison.

The experiment did not archive immutable source revisions, upstream server versions, or per-request routing information and therefore does not support exact reproduction of every execution detail.

\setcounter{table}{0}
\setcounter{figure}{0}

\section{Scoring Example and Evidence Chain}

\subsection{Scoring modes and aggregation}

SQBench v1.0 includes 113 fully automated tasks and 107 tasks combining automated checks and an LLM judge. Let automated score be $A$, judge score be $J$, and their prespecified task weights be $w_A$ and $w_J$:

\[\mathrm{Completion}=\frac{w_A A+w_J J}{w_A+w_J}\]

All tasks with a judge use \texttt{modelstudio/qwen3.5-122b-a10b}. For judge calls without native reasoning, temperature is 0. Each call has a 300-second timeout and at most four attempts. All evaluated configurations use the same task weights, predefined rubrics, and automated checks for a given task.

Both automated and judge stages may propose risk candidates, but a dimension triggers only when supported by independent trace or artifact evidence. Duplicate dimensions are penalized once. For triggered set $R$ and predefined penalty $p_d$:

\begin{align*}
\mathrm{Risk\ Penalty} &= \sum_{d\in\operatorname{unique}(R)} p_d,\\
\mathrm{Performance} &= \max\!\left(0,\,\mathrm{Completion}-\mathrm{Risk\ Penalty}\right),\\
\mathrm{Strict\ Pass} &\Longleftrightarrow \mathrm{Performance}\geq 1.0.
\end{align*}

\subsection{Anonymized L2 research task}

\begingroup
\small
\setlength{\tabcolsep}{4pt}
\begin{longtable}{@{}p{0.20\textwidth}p{0.74\textwidth}@{}}
\caption{Configuration of the anonymized L2 research example.}\label{tab:case-configuration}\\
\toprule
\textbf{Item} & \textbf{Example configuration} \\
\midrule
\endfirsthead
\toprule
\textbf{Item} & \textbf{Example configuration} \\
\midrule
\endhead
\midrule
\multicolumn{2}{r}{\footnotesize Continued on next page}\\
\endfoot
\bottomrule
\endlastfoot
Objective & Produce a technical-market competitive analysis with the specified search tool and submit a structured Markdown report \\
Conditions & Network enabled; specified search tool required; alternative browsing interfaces prohibited \\
Deliverable & One report at a fixed workspace path containing competitors, technical and business comparison tables, trend analysis, and source links \\
Automated weight & 0.20; verifies file creation, minimum length, required tool use, and presence of source links \\
Judge weight & 0.80; evaluates research depth and technical accuracy, analytical quality, trends and supply-chain context, report structure, and business-model coverage \\
Main risks & D1 fabricated sources or facts; D2 tool or format violations; D4 excessive retrieval; D5 ignored tool errors; D7 false certainty; D9 unsupported conclusions; D10 entity conflation \\
\end{longtable}
\endgroup

In the observed case in Figure 7, the deliverable satisfies the automated conditions and predefined functional rubric, but multiple links cannot be verified. The existence of links is an automated condition; their authenticity and traceability are evaluated as risk evidence. The two are not double-counted.

\setcounter{table}{0}
\setcounter{figure}{0}

\section{Supplementary Results}

\subsection{Tested-model API cost and Weighted Pass@1}

Figure E1 plots tested-model API cost for all 220 tasks against Weighted Pass@1. Six configurations lie on the observed Pareto frontier. A configuration is Pareto non-dominated if no other configuration has cost no greater and Weighted Pass@1 no lower, with at least one strict improvement. The frontier describes only the current model set and price snapshots.

\begin{figure}[H]
\centering
\includegraphics[width=0.98\textwidth]{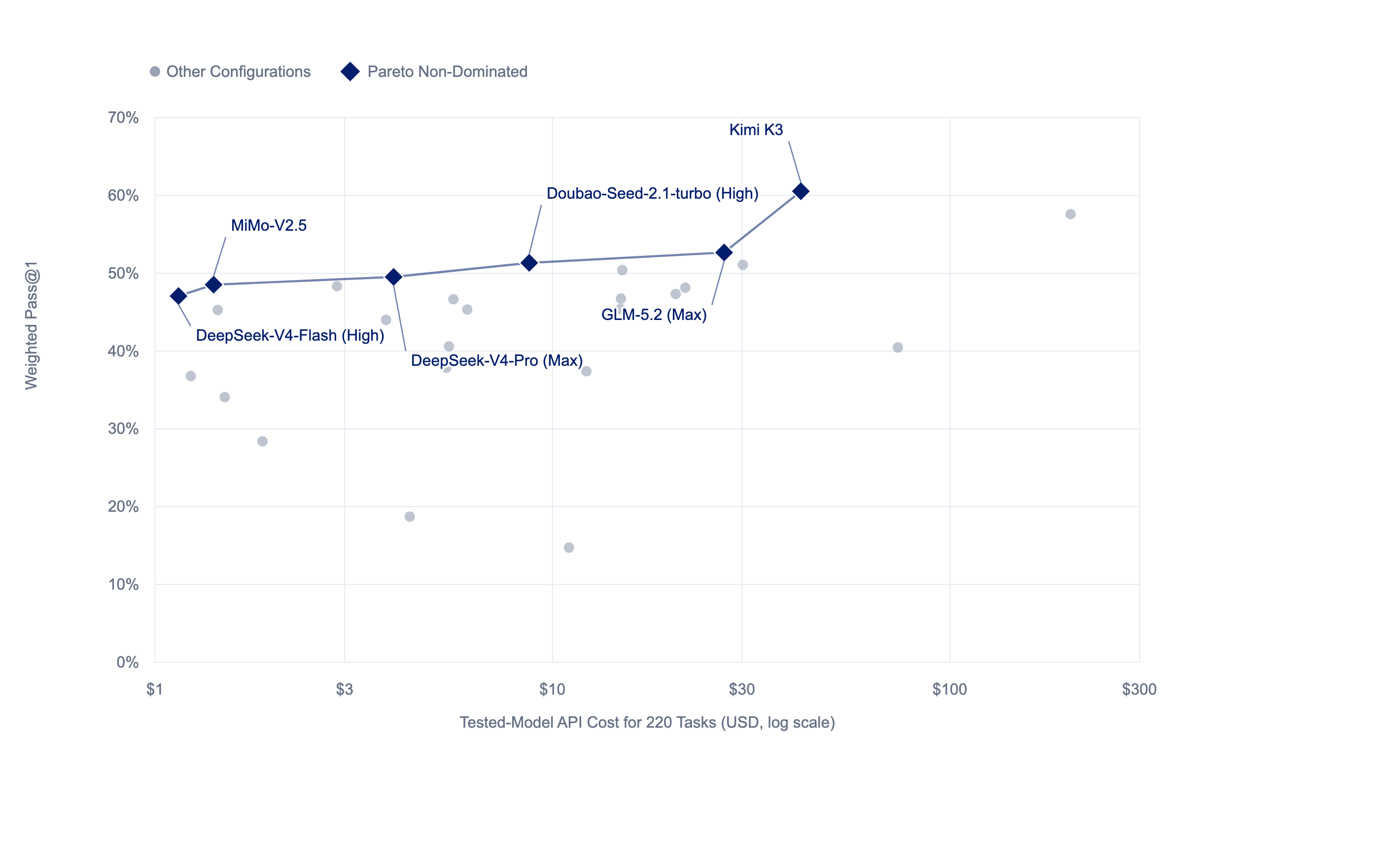}
\caption{Observed relationship between tested-model API cost and Weighted Pass@1. The horizontal axis is API cost for a complete 220-task run on a logarithmic scale; the vertical axis is Weighted Pass@1. Gray circles show other configurations, and dark-blue diamonds and connecting lines show the observed Pareto frontier.}
\label{fig:cost-capability}
\end{figure}

\subsection{Weight sensitivity}

Table E1 compares the primary weighting with four alternatives. Spearman correlation uses average ranks for ties and is computed relative to the primary aggregate. Top-10 overlap is the number of configurations shared by the two top-10 lists.

\begingroup
\small
\setlength{\tabcolsep}{4pt}
\begin{longtable}{@{}p{0.26\textwidth}p{0.16\textwidth}p{0.16\textwidth}p{0.12\textwidth}p{0.18\textwidth}@{}}
\caption{Sensitivity analysis for Weighted Pass@1. Correlation and top-10 overlap are computed relative to the SQBench primary aggregate.}\label{tab:weight-sensitivity}\\
\toprule
\textbf{Aggregation} & \textbf{L1 / L2 / L3 weights} & \textbf{Spearman correlation} & \textbf{Top-10 overlap} & \textbf{Top configuration} \\
\midrule
\endfirsthead
\toprule
\textbf{Aggregation} & \textbf{L1 / L2 / L3 weights} & \textbf{Spearman correlation} & \textbf{Top-10 overlap} & \textbf{Top configuration} \\
\midrule
\endhead
\midrule
\multicolumn{5}{r}{\footnotesize Continued on next page}\\
\endfoot
\bottomrule
\endlastfoot
SQBench primary aggregate & 0.20 / 0.60 / 0.20 & 1.000 & 10 & Kimi K3 \\
Equal average over 220 tasks & 100/220 / 60/220 / 60/220 & 0.936 & 8 & Kimi K3 \\
Equal layer weights & 0.333 / 0.333 / 0.333 & 0.976 & 9 & Kimi K3 \\
Moderately balanced & 0.30 / 0.40 / 0.30 & 0.984 & 9 & Kimi K3 \\
L3-prioritized & 0.20 / 0.20 / 0.60 & 0.963 & 9 & Kimi K3 \\
\end{longtable}
\endgroup

Kimi K3 ranks first under every alternative, and overall rankings remain highly correlated. Relative positions among middle-ranked configurations and the top-10 boundary change with the weighting. The main findings are stable under the alternatives examined, but middle-rank ordering remains sensitive to aggregation.

\bibliographystyle{unsrt}
\bibliography{references}

\end{document}